\documentclass[%
 reprint,
superscriptaddress,
 amsmath,amssymb,
 aps,
]{revtex4-2}

\usepackage{graphicx}
\usepackage{dcolumn}
\usepackage{bm}

\begin{document}

\preprint{APS/123-QED}

\title{Optimal Memory Encoding Through Fluctuation–Response Structure
}

\author{Lianxiang Cui}
\email{lxcui@g.ecc.u-tokyo.ac.jp}
\affiliation{International Research Center for Neurointelligence, University of Tokyo Institutes for Advanced Study, University of Tokyo, Bunkyo-ku, Tokyo 113-0033, Japan}

\author{Kohei Nakajima}%
\affiliation{Graduate School of Information Science and Technology, University of Tokyo, Bunkyo-ku, Tokyo 113-0033, Japan}
\affiliation{International Research Center for Neurointelligence, University of Tokyo Institutes for Advanced Study, University of Tokyo, Bunkyo-ku, Tokyo 113-0033, Japan}

\author{Kazuyuki Aihara}
\affiliation{International Research Center for Neurointelligence, University of Tokyo Institutes for Advanced Study, University of Tokyo, Bunkyo-ku, Tokyo 113-0033, Japan}

\date{\today}

\begin{abstract}
Physical reservoir computing exploits the intrinsic dynamics of physical systems for information processing, while keeping the internal dynamics fixed and training only linear readouts; yet the role of input encoding remains poorly understood.
We show that optimal input encoding is a geometric problem governed by the system’s fluctuation–response structure. 
By measuring steady-state fluctuations and linear response, we derive an analytical criterion for the input direction that maximizes task-specific linear memory under a fixed power constraint, termed Response-based Optimal Memory Encoding (ROME).
Backpropagation-based encoder optimization is shown to be equivalent to ROME, revealing a trade-off between task-dependent feature mixing and intrinsic noise.
We apply ROME to various reservoir platforms, including spin-wave waveguides and spiking neural networks, demonstrating effective encoder design across physical and neuromorphic reservoirs, even in non-differentiable systems.
\end{abstract}

\maketitle


Reservoir computing (RC) is a widely used 
information processing paradigm that exploits nonlinear dynamics as its computational resource
and can be viewed as a class of recurrent neural networks with fixed random internal connectivity \cite{nakajima2021book, yan2024emerging, gonon2019reservoir}.
The reservoir’s high-dimensional internal dynamics transform input signals into rich state trajectories, while only a linear readout is usually trained to perform the target task.
As a result, RC avoids large-scale parameter optimization, which makes learning efficient and quick.
At the same time, the use of fixed random internal networks has motivated implementations ranging from abstract neural models to physical systems, which is termed physical reservoir computing (PRC) \cite{Nakajima_PRC_2020, TANAKA2019}. These systems include biological substrates \cite{Matteo2021, pnas2023, cai2023brain, ushio2023computational}, quantum \cite{quantumPRC_2017, martinez2021dynamical, nakajima2019boosting, mujal2021opportunities}, photonic \cite{van2017advances, vandoorne2014experimental,larger2017high, paquot2012optoelectronic}, spintronic \cite{torrejon2017neuromorphic, nakajima_2019_spin} and memristive systems \cite{du2017reservoir, moon2019temporal, zhong2021dynamic}.

In addition to the internal dynamics and the readout layer, the input layer plays an equally critical role in determining reservoir performance, as it directly controls how much information is encoded into the system and how it is represented by the system dynamics.
However, in the standard RC formulation, input connections are typically chosen at random, leaving open the question of whether an optimal input coupling exists that maximizes performance for a given task and system.
Recent studies have shown that optimizing the input matrix via backpropagation (BP) or gradient-based approach can indeed improve RC performance \cite{THIEDE201923, OZTURK2020215, Liu_nakajima2025, Tomioka_nakajima2025, tanaka2022continuum}. However, such approaches treat input optimization purely as a machine learning problem.
More fundamentally, different tasks exhibit markedly different performance across different dynamical systems, raising a deeper question: what determines the compatibility between a given task and a given reservoir \cite{tsunegi2023information, kurikawa2025fluctuation}?
Purely data-driven optimization provides limited insight into this issue.
Addressing these questions requires treating the reservoir as a noisy, high-dimensional dynamical system, in which the encoding and retention of information are constrained by the system’s intrinsic dynamical structure.

In this letter, we show that optimal input encoding and system–task compatibility are governed by the geometric structure induced by the system’s fluctuation and response.
Specifically, the ability of a system to encode and memorize task-relevant information is constrained by the interplay between its steady-state fluctuations and its response to external inputs.
Within this framework, a task is feasible only if its relevant input features can be aligned with low-noise, long-lived dynamical modes that exhibit strong response relative to intrinsic fluctuations.

This perspective leads to a principled criterion for optimal input encoding, which we term Response-based Optimal Memory Encoding (ROME).
By measuring the steady-state fluctuations and linear response of the system, ROME identifies the input direction that maximizes task-specific linear memory under a fixed input power constraint.
We further show that BP-based optimization of the input matrix is equivalent to this response-based criterion, in the sense that BP searches for input directions that optimally trade off task-dependent feature mixing against the reservoir’s intrinsic noise structure.
Finally, we demonstrate ROME in a spin-wave physical reservoir and a heterogeneous excitatory–inhibitory (E/I) spiking neural network (SNN) reservoir, highlighting the potential of this approach for PRC and non-differentiable system.

\textit{ROME}---Consider a general noisy dynamical system used as a reservoir, whose dynamics are described by
\begin{equation}
\dot{\mathbf x}(t)=\mathbf f(\mathbf x(t)) + G\,\mathbf u(t) + \boldsymbol{\xi}(t),
\qquad 
\langle \boldsymbol{\xi}(t)\rangle=0,
\label{eq:reservoir}
\end{equation}
where $\mathbf x(t)\in\mathbb R^N$ is the reservoir state, $\mathbf f:\mathbb R^N\to\mathbb R^N$ is an arbitrary (possibly nonlinear) drift term, $\mathbf u(t)\in\mathbb R^m$ is the input signal, $G$ is the input matrix, and $\boldsymbol{\xi}(t)$ represents intrinsic noise. We assume that in the absence of input ($\mathbf u\equiv 0$), the system admits a unique stationary distribution $\rho_0$ and the steady-state covariance $\Sigma_{\rm ref}=\mathbb E_{\rho_0}\!\left[\mathbf x(t)\mathbf x(t)^\top\right]$
exists, quantifying intrinsic fluctuations around the zero-input reference working point. 
In the linear-response regime, the first-order response is 
$\delta\langle \mathbf x(t)\rangle
=
\int_0^\infty R(\tau)\,G\,\delta\mathbf u(t-\tau)\,\mathrm d\tau,$
where $R(\tau)$ is the response kernel, characterizing the reservoir’s intrinsic response structure at the reference working point.

To assess how the reservoir encodes input signals, we adopt the generalized high-dimensional memory function (MF) as our metric, defined as the ability of a linear readout $\mathbf y(t)$ to reconstruct the delayed input $\mathbf u(t-\tau)$, given by
\begin{equation}
\mathrm{MF}(\tau;G)
\equiv
\operatorname{Tr}\!\Big(
\Sigma_{xu}(\tau)\,
\Sigma_{\rm sig}^{-1}\,
\Sigma_{ux}(\tau)\,
\Sigma_{\rm ref}^{-1}
\Big),
\label{eq:MF}
\end{equation}
where
$
\Sigma_{xu}(\tau)
=
\mathbb E\!\left[\mathbf x(t)\mathbf u(t-\tau)^\top\right]
=
\int_0^\infty
R(s)\,G\,\Sigma_{uu}(s-\tau)\,\mathrm ds
$
is the input–state cross-covariance and $\Sigma_{\rm sig}$ denotes the covariance of the input signal.
For a scalar input ($m=1$), this expression reduces to the standard linear memory function used in RC \cite{jaeger2001short, dambre2012information, guan2025noise}, $\text{MF}(\tau)=\mathrm{corr}^2\!\big(y(t),u(t-\tau)\big)$ (see Supplementary Material Sec.~I).
For white-noise input, 
the memory function reduces to the explicit fluctuation–response form
\begin{equation}
\mathrm{MF}(\tau;G)
=
\operatorname{Tr}\!\Big(
R(\tau)\,\tilde G\,\tilde G^\top R(\tau)^\top\,
\Sigma_{\rm ref}^{-1}
\Big),
\end{equation}
where $\tilde G = G\,\Sigma_{\rm sig}^{1/2}$ denotes the whitened input matrix.
In the scalar input case, this expression simplifies to
$
\mathrm{MF}(\tau;G)
=
\big\|R(\tau)\,\tilde G\big\|_{\Sigma_{\rm ref}^{-1}}^2,
$
which corresponds to the squared linear response of the reservoir measured in the fluctuation metric.

For generality, we introduce a nonnegative task-weight spectrum $w(\tau)\ge 0$ to represent mixtures of different temporal delays. The corresponding task-weighted memory objective is defined as $\mathcal J_w(G)=\int_{0}^{\infty} w(\tau)\,\mathrm{MF}(\tau;G)\,\mathrm d\tau$. When $w(\tau)=\delta(\tau-\tau_0)$, this reduces to the single-delay memory function $\mathrm{MF}(\tau_0;G)$.
Maximizing the memory objective under a fixed input power constraint now becomes a purely geometric optimization problem.  
For a fixed input power $P$, which corresponds to the parameter “input scale” in RC, the optimization problem can be written as
\begin{equation}
\mathcal J_w^\star=
\max_{\tilde G}\;
\operatorname{Tr}\!\Big(
\tilde G^\top M_w \tilde G
\Big)
\quad
\text{s.t.}
\|\tilde G\|_F^2 \leq P,
\end{equation}
where $M_w=\int_0^\infty
w(\tau)\, 
R(\tau)^\top\Sigma_{\rm ref}^{-1}R(\tau)\mathrm d\tau$.
This is a standard Rayleigh–Ritz optimization whose solution is given by the dominant eigenmodes of the symmetric operator $M_w$. The optimal input matrix is $\tilde G^\star=\sqrt{P}\,v_1 \,e_1^\top$, where $v_1$ denotes the largest eigenvector of $M_w$, and $e_1$ is the first standard basis vector in the input space $\mathbb{R}^m$. 
If $r$ active directions are required, one selects the $r$ leading eigenvectors $v_i$ and constructs the encoder as $G=\sum_{i=1}^r \sqrt{p_i}\, v_i e_i^\top$, where $p_i \ge 0$ and $\sum_{i=1}^r p_i=P$.

For an arbitrary dynamical system, ROME identifies the optimal encoding direction for linear memory by measuring its steady-state fluctuations and linear response.
This approach encodes task-relevant features into the reservoir’s slow dynamical modes, thereby minimizing the impact of intrinsic noise on the task representation.
Consequently, a task is feasible only if its relevant input features can be aligned with dynamical directions that are long-lived, low-noise, and exhibit strong response relative to intrinsic fluctuations.
More generally, the memory objective can be replaced by a task representation, allowing the optimization of all linear memory tasks, which correspond to weighted combinations of linear memory across timescales (see Supplementary Material Sec.~II).

\textit{Equivalence to Backpropagation}---We now establish the equivalence between ROME and BP-based optimization of the input matrix.
With the reservoir dynamics fixed, for any given input matrix $G$, the output is always constructed using the optimal linear readout obtained from ordinary least-squares (OLS) regression.
The readout $W_{\rm out}$ is therefore not trained by BP but calculated analytically, thereby removing the influence of the readout layer on the results.
Consequently, BP is used solely to optimize the input matrix $G$ through the prediction error under the optimal linear readout.

Without loss of generality, we consider the single-delay memory reconstruction task, which predicts the past input $\mathbf u(t-\tau)$ from the current reservoir state $\mathbf x(t)$ using a linear readout,
$\mathbf y(t)=W_{\rm out}\mathbf x(t)$.
The objective is to minimize the mean-squared error
$
\mathcal E(\tau;G)
=
\mathbb E\big\|
\mathbf y(t)-\mathbf u(t-\tau)
\big\|_2^2 .
$
For a given $G$, the optimal OLS readout matrix is
$
W_{\rm out}^\star(\tau)
=
\Sigma_{ux}(\tau)\,\Sigma_x^{-1},
$
where $\Sigma_x=\mathbb E[\mathbf x(t)\mathbf x(t)^\top]$.
Substituting this solution yields the minimal prediction error
\begin{equation}
\mathcal E^\star(\tau;G)
\propto
-
\operatorname{Tr}\!\Big(
\Sigma_{xu}(\tau)\,
\Sigma_{ux}(\tau)\,
\Sigma_x^{-1}
\Big).
\end{equation}
Since the input covariance $\Sigma_{\rm sig}$ is fixed and independent of $G$, and at the chosen reference working point $\Sigma_{\rm ref}=\Sigma_x$, minimizing the BP objective $\mathcal E^\star(\tau;G)$ is monotonically equivalent to maximizing the memory function $\mathrm{MF}(\tau;G)$ defined in Eq.~\eqref{eq:MF}.
Thus, BP converges to the same optimal input encoding directions as ROME.

\begin{figure*}[htbp]
  \centering
  \includegraphics[width=0.99\textwidth]{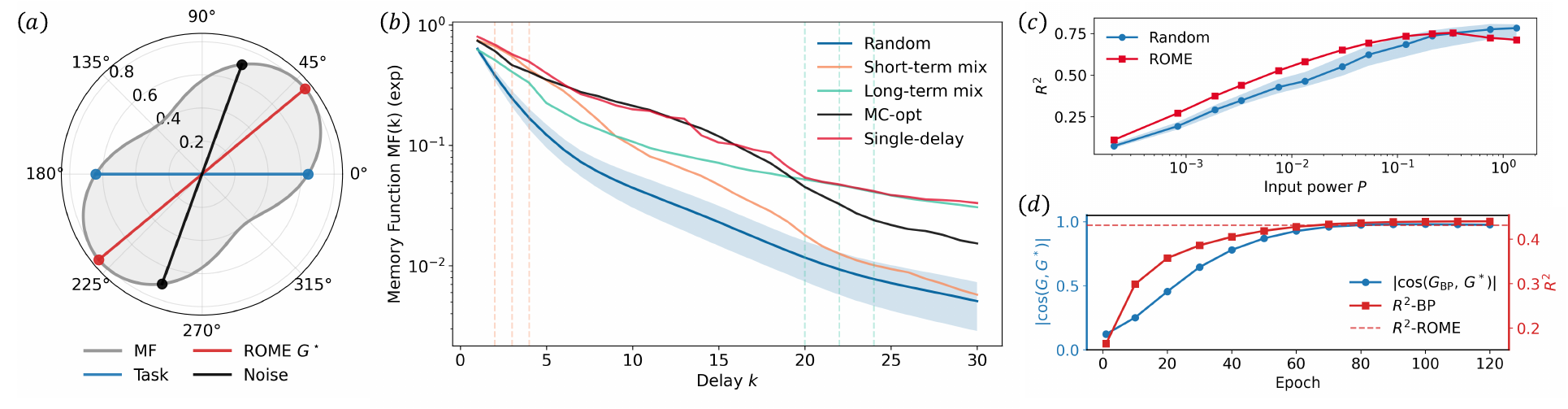}
  \caption{
  Numerical demonstrations of ROME. 
  (a) Polar plot of the MF for a linear reservoir, evaluated in the plane spanned by the task-only direction (blue) and the ROME-optimal direction (red); gray curve: MF landscape; black line: noise-minimizing direction projected onto this plane.
  (b) MF versus delay $k$ for a linear reservoir with different linear memory tasks. Blue curve and shaded band: mean$\pm$s.d. over random encoders; colored curves: short-term ($k=2,3,4$; orange) and long-term ($k=20,22,24$; green) ROME encoders, MC-optimal encoder (black, $\mathrm{MC}=\sum_k \mathrm{MF}(k)$), and single-delay–optimal encoder (red). 
  (c) NARMA10 performance $R^2$ versus input power $P$ for a nonlinear ESN. Blue circles and shaded band: mean$\pm$s.d. over random encoders; Red squares: ROME encoder. 
  (d) Training evolution for the BP encoder in a linear reservoir (single-delay task). Blue circles: alignment $|\cos(G_{\rm BP},G^*)|$ between learned input direction and ROME direction; red squares: BP performance $R^2$; dashed line: $R^2$ of the ROME direction $G^*$. 
  }
  \label{fig:1}
\end{figure*}

This equivalence provides a deeper understanding of neural network training.
The role of the input layer is to align task-relevant data features with the intrinsic structure of the dynamical system, thereby maximizing the amount of information that can be encoded in the network state.
While we illustrate this mechanism using a single-delay memory task, different tasks correspond to different feature structures, yet the underlying principle remains unchanged.
Accordingly, whether a given neural network can perform a particular task is determined by the alignment between the task’s feature directions and the system’s internal dynamical structure.

\textit{Numerical Results}---We first illustrate the geometric structure underlying ROME using a linear reservoir with $\mathbf f(\mathbf x)=W\mathbf x$ and a single-delay task. 
Figure~\ref{fig:1}(a) shows the MF landscape under a fixed power constraint, restricted to the plane spanned by the task-only direction and the ROME direction. 
The optimal encoder corresponds to the principal axis of this landscape. 
Notably, it does not coincide with the task-only direction, which maximizes signal gain, nor with the noise-minimizing direction, which suppresses intrinsic fluctuations. 
Instead, the ROME direction emerges from their geometric competition.

Figure~\ref{fig:1}(b) compares $\mathrm{MF}(k)$ obtained with different encoder constructions. 
Relative to random encoders, all ROME-designed encoders yield a pronounced enhancement of memory. 
The single-delay optimum provides an empirical upper envelope, indicating the best achievable memory at each delay when the encoder is tuned to that delay alone. 
By contrast, the MC-optimal encoder spreads memory across delays and therefore does not concentrate strongly on any specific $k$. 
The mixed-delay ROME encoders selectively amplify memory at the prescribed delays (orange for $k=2,3,4$; green for $k=20,22,24$), outperforming the MC-optimal encoder in their targeted regimes and exposing the trade-off between task specialization and global memory allocation.


To assess the performance of ROME in nonlinear reservoirs and on complex tasks, we apply it to an echo-state network (ESN) \cite{jaeger2001echo,Jaeger_2004_esn} solving the NARMA10 benchmark task. 
The reservoir dynamics are given by 
$\mathbf x_{t+1}=\tanh(W\mathbf x_t+G\,u_t+\boldsymbol{\xi}_t)$ with spectral radius $\rho(W)<1$, 
while the target output follows the NARMA10 dynamics
\begin{equation}
y_{t+1}
=
0.3\,y_t
+
0.05\,y_t
\left(\sum_{i=0}^{9} y_{t-i}\right)
+
1.5\,u_{t-9}u_t
+
0.1 ,
\end{equation}
where the input $u_t$ is drawn from a positive-valued uniform distribution with a nonzero mean.
As shown in Fig.~\ref{fig:1}(c), in the low-input-power regime, the ROME direction consistently outperforms random encoders, demonstrating its effectiveness in nonlinear reservoirs. 
When the performance $R^2$ becomes high and the input power $P$ is further increased, random encoders eventually surpass ROME. 
This behavior can be attributed to two factors: 
(1) the NARMA10 task requires nonlinear memory \cite{Kubota2021}, whereas the current ROME optimizes only linear memory contributions, limiting the achievable improvement in $R^2$; 
(2) increasing the input power drives the reservoir away from the chosen reference working point, leading to changes in the internal fluctuation–response structure that are not captured by the fixed-metric optimization.

In Fig.~\ref{fig:1}(d), we demonstrate the equivalence between ROME and BP-based encoding optimization in a linear reservoir performing a single-delay task.
During training, the input direction $G_{\rm BP}$ learned by BP progressively aligns with the ROME direction $G^*$, while the performance $R^2$ converges to the ROME baseline. 
This demonstrates the equivalence between BP and ROME encoder, showing that BP training also finds the optimal fluctuation–response encoding direction.

\begin{figure}[tbp]
  \centering
  \includegraphics[width=\columnwidth]{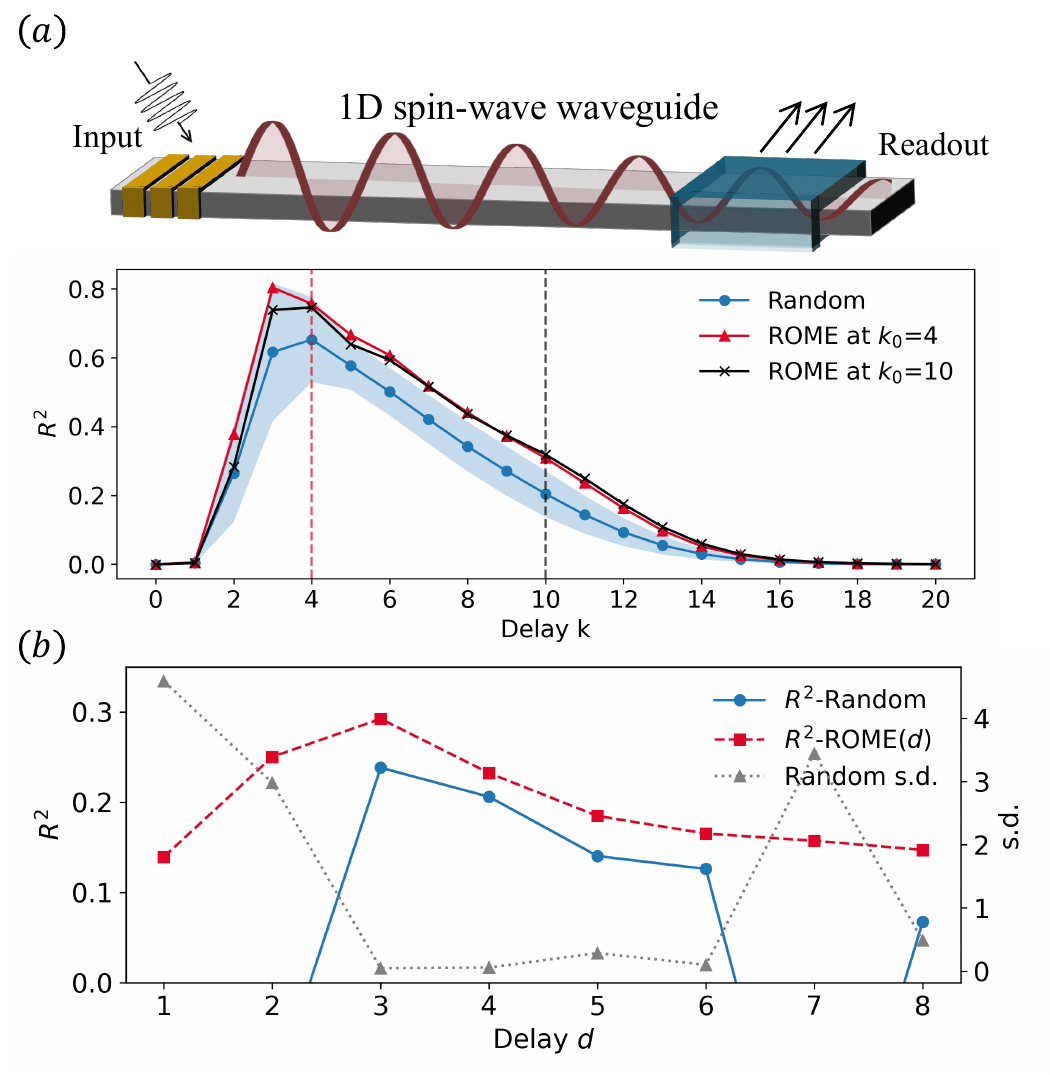}
  \caption{PRC demonstrations of ROME. (a) Schematic illustration of the spin-wave physical reservoir and the performance $R^2$ versus delay $k$ on a single-delay memory task. Blue curve and shaded band: mean $\pm$ s.d. over random encoders; red and black: ROME encoders optimized for target delays $k_0$. (b) Performance of the heterogeneous E/I SNN neuromorphic PRC systems on a single-delay memory task. Red dashed line: ROME; blue line: mean over random encoders; gray dotted lines: standard deviation of random encoders.}
  \label{fig:2}
\end{figure}

\textit{Physical reservoir computing}---As ROME relies only on measurable fluctuation and response statistics, without requiring detailed knowledge of the underlying dynamics, it is naturally suited for the training of physical reservoir computing (PRC). 
Here, we demonstrate ROME on two PRC platforms: a one-dimensional noisy traveling spin-wave waveguide \cite{collet2017spin, spinwave_PRC, spinwave_PRC2, nikolaev2024resonant, spinwave_waveguide, imai2025gradient}, shown in Fig.~\ref{fig:2}(a), and a heterogeneous E/I SNN \cite{MAASS2004593, nicola2017supervised, sakemi2023sparse, xue2013computational}. 
In contrast to abstract recurrent neural networks, information processing in these systems is governed by their underlying physical or neuromorphic dynamics.

We first consider the 1D traveling spin-wave waveguide in a single-delay memory task. 
The microscopic dynamics of the magnetization field is governed by the Landau--Lifshitz--Gilbert (LLG) equation \cite{lakshmanan2011fascinating, nakatani1989direct, chumak2015magnon}. 
Under a static bias field and weak excitation, the dynamics can be reduced to an effective linear spin-wave description around a stable operating point (see Supplementary Material Sec.~III for details).
By introducing a complex spin-wave envelope field $\psi(x,t)$ for the one-dimensional waveguide and expanding the dispersion relation around a carrier wavenumber, one obtains the effective linear envelope equation \cite{Kalinikos_1986, Rezende2009}:
\begin{equation}
    \partial_t \psi
=
(-\Gamma+i\Omega_0)\psi
- v_g\,\partial_x \psi
+ iD\,\partial_x^2 \psi
+ b(x)\,u(t)
+ \xi(x,t),
\end{equation}
where $\Omega_0$ denotes the carrier angular frequency at the operating point, $v_g$ is the group velocity, $D$ the dispersion curvature, $\Gamma$ the damping rate, $b(x)$ the spatial input profile, and $\xi(x,t)$ additive noise.

We numerically integrate this envelope equation, and the reservoir state is given by the real and imaginary parts of the spin-wave envelope $\psi$. To reflect realistic experimental situations, the input is injected through a spatially localized profile $b(x)$ over the first $w_{\rm in}$ region of the waveguide, while readout features are constructed from local measurements of the spin-wave amplitude within a spatial window of width $w_{\rm out}$. 

By measuring the steady-state fluctuations of the spin-wave field and the linear response of the system to impulsive perturbations at a target delay $k_0$, ROME identifies the optimal spatial input profile $b^\star(x)$, corresponding to an effective injection antenna. 
As shown in Fig.~\ref{fig:2}(a), ROME encodings consistently outperform random input weights, and at a given delay $k_0$ the corresponding ROME-optimized encoding yields the best performance.

We next consider the neuromorphic reservoir composed of a SNN with leaky integrate-and-fire (LIF) neurons (see Supplementary Material Sec.~IV for details). 
To better reflect biologically plausible circuitry, the network incorporates neuronal heterogeneity and distinct excitatory and inhibitory populations.  
The input signal and reservoir readout act on distinct subsets of neurons, mimicking localized stimulation and partial observation.
In this setting, the reservoir state is defined by filtered spike trains, and a linear readout is trained to perform a single-delay memory task.

Similarly, by measuring the steady-state fluctuations and the linear response, ROME determines an optimal input encoding matrix for this heterogeneous E/I SNN reservoir. 
Due to the partial input injection and readout, together with complex internal dynamics and spike-based variability, the network exhibits highly unstable performance under random input encodings, often yielding negative $R^2$ values, as shown in Fig.~\ref{fig:2}(b). 
Moreover, the standard deviation of the $R^2$ increases markedly with the number of samples, indicating strong sensitivity to the choice of the input matrix. 
In contrast, ROME identifies a stable optimal encoding that systematically improves task performance. 
Notably, owing to the discrete spiking dynamics, SNNs are intrinsically non-differentiable, rendering BP-based optimization methods inapplicable. 
ROME therefore provides a practical and physically grounded approach for input design in spiking neural reservoirs.

\textit{Conclusion and discussion}---In this letter, we introduced ROME, a fluctuation–response–based framework for optimizing input encoding in RC systems. By exploiting only measurable fluctuations and response statistics at a chosen working point, ROME identifies input directions that optimally balance task relevance and noise suppression. 
ROME provides a unified interpretation of encoder optimization as a geometric problem, where useful computation emerges from selectively exciting long-lived, low-noise dynamical directions. Moreover, the equivalence between ROME and BP-based optimization reveals a common underlying principle of learning: both methods effectively seek an optimal trade-off between task-dependent feature mixing and intrinsic noise.

For realistic physical systems, detailed dynamical models are often unknown, inaccessible, or intrinsically non-differentiable, making conventional gradient-based optimization impractical. 
These considerations render ROME particularly well suited for input optimization in PRC, as illustrated in both a spin-wave waveguide and a heterogeneous E/I SNN, and naturally extend its relevance to the development of brain-inspired computational architectures.

From the perspective of the fluctuation–dissipation theorem, equilibrium systems obey a proportionality between spontaneous fluctuations and linear response, fixed by a scalar temperature \cite{Einstein_FDT, FDT1951, kubo1966fluctuation}. 
However, driven or nonequilibrium systems generally exhibit anisotropic and direction-dependent fluctuation–response structure, allowing certain dynamical directions to display strong linear response relative to intrinsic fluctuations over task-relevant timescales. ROME can thus be viewed as selecting input directions that minimize an effective, direction- and timescale-dependent fluctuation–response ratios. In this sense, the optimal encoding direction can be regarded as computationally “cold.” This “coldness” should not, however, be interpreted as a thermodynamic temperature, but rather as an operational measure of noise relative to controllability.

The present formulation of ROME is based on a fixed reference working point and optimizes linear memory contributions; extensions to self-consistent working points and nonlinear memory constitute natural directions for future work.

\textit{Acknowledgments}---This research was supported by Moonshot R\&D (Grant No. JPMJMS2021), the Institute of AI and Beyond of UTokyo, the International Research Center for Neurointelligence (WPI-IRCN) at The University of Tokyo Institutes for Advanced Study (UTIAS), Council for Science, Technology and Innovation (CSTI), the Cross-ministerial Strategic Innovation Promotion Program (SIP), the 3rd period of SIP (Grant No. JPJ012207), a project (Grant No. JPNP14004) commissioned by the New Energy and Industrial Technology Development Organization (NEDO), and JSPS KAKENHI (Grant Number 25H01134).



\bibliography{apssamp}

@book{nakajima2021book,
  title={Reservoir computing},
  author={Nakajima, Kohei and Fischer, Ingo},
  year={2021},
  publisher={Springer}
}

@article{TANAKA2019,
title = {Recent advances in physical reservoir computing: A review},
journal = {Neural Networks},
volume = {115},
pages = {100-123},
year = {2019},
issn = {0893-6080},
doi = {https://doi.org/10.1016/j.neunet.2019.03.005},
url = {https://www.sciencedirect.com/science/article/pii/S0893608019300784},
author = {Gouhei Tanaka and Toshiyuki Yamane and Jean Benoit Héroux and Ryosho Nakane and Naoki Kanazawa and Seiji Takeda and Hidetoshi Numata and Daiju Nakano and Akira Hirose}
}

@article{Nakajima_PRC_2020,
doi = {10.35848/1347-4065/ab8d4f},
url = {https://doi.org/10.35848/1347-4065/ab8d4f},
year = {2020},
month = {may},
publisher = {IOP Publishing},
volume = {59},
number = {6},
pages = {060501},
author = {Nakajima, Kohei},
title = {Physical reservoir computing—an introductory perspective},
journal = {Japanese Journal of Applied Physics}
}

@article{yan2024emerging,
  title={Emerging opportunities and challenges for the future of reservoir computing},
  author={Yan, Min and Huang, Can and Bienstman, Peter and Tino, Peter and Lin, Wei and Sun, Jie},
  journal={Nature Communications},
  volume={15},
  number={1},
  pages={2056},
  year={2024},
  publisher={Nature Publishing Group UK London}
}

@article{van2017advances,
  title={Advances in photonic reservoir computing},
  author={Van der Sande, Guy and Brunner, Daniel and Soriano, Miguel C},
  journal={Nanophotonics},
  volume={6},
  number={3},
  pages={561--576},
  year={2017},
  publisher={De Gruyter}
}

@article{vandoorne2014experimental,
  title={Experimental demonstration of reservoir computing on a silicon photonics chip},
  author={Vandoorne, Kristof and Mechet, Pauline and Van Vaerenbergh, Thomas and Fiers, Martin and Morthier, Geert and Verstraeten, David and Schrauwen, Benjamin and Dambre, Joni and Bienstman, Peter},
  journal={Nature communications},
  volume={5},
  number={1},
  pages={3541},
  year={2014},
  publisher={Nature Publishing Group UK London}
}

@article{larger2017high,
  title={High-speed photonic reservoir computing using a time-delay-based architecture: Million words per second classification},
  author={Larger, Laurent and Bayl{\'o}n-Fuentes, Antonio and Martinenghi, Romain and Udaltsov, Vladimir S and Chembo, Yanne K and Jacquot, Maxime},
  journal={Physical Review X},
  volume={7},
  number={1},
  pages={011015},
  year={2017},
  publisher={APS}
}

@article{nakajima_2019_spin,
    author = {Tsunegi, Sumito and Taniguchi, Tomohiro and Nakajima, Kohei and Miwa, Shinji and Yakushiji, Kay and Fukushima, Akio and Yuasa, Shinji and Kubota, Hitoshi},
    title = {Physical reservoir computing based on spin torque oscillator with forced synchronization},
    journal = {Applied Physics Letters},
    volume = {114},
    number = {16},
    pages = {164101},
    year = {2019},
    month = {04},
    issn = {0003-6951},
    doi = {10.1063/1.5081797},
    url = {https://doi.org/10.1063/1.5081797},
}

@article{cai2023brain,
  title={Brain organoid reservoir computing for artificial intelligence},
  author={Cai, Hongwei and Ao, Zheng and Tian, Chunhui and Wu, Zhuhao and Liu, Hongcheng and Tchieu, Jason and Gu, Mingxia and Mackie, Ken and Guo, Feng},
  journal={Nature Electronics},
  volume={6},
  number={12},
  pages={1032--1039},
  year={2023},
  publisher={Nature Publishing Group UK London}
}

@article{gonon2019reservoir,
  title={Reservoir computing universality with stochastic inputs},
  author={Gonon, Lukas and Ortega, Juan-Pablo},
  journal={IEEE transactions on neural networks and learning systems},
  volume={31},
  number={1},
  pages={100--112},
  year={2019},
  publisher={IEEE}
}

@article{lakshmanan2011fascinating,
  title={The fascinating world of the Landau--Lifshitz--Gilbert equation: an overview},
  author={Lakshmanan, Muthusamy},
  journal={Philosophical Transactions of the Royal Society A: Mathematical, Physical and Engineering Sciences},
  volume={369},
  number={1939},
  pages={1280--1300},
  year={2011},
  publisher={The Royal Society Publishing}
}

@article{nakatani1989direct,
  title={Direct solution of the Landau-Lifshitz-Gilbert equation for micromagnetics},
  author={Nakatani, Yoshinobu and Uesaka, Yasutaro and Hayashi, Nobuo},
  journal={Japanese Journal of Applied Physics},
  volume={28},
  number={12R},
  pages={2485},
  year={1989},
  publisher={IOP Publishing}
}

@article{spinwave_PRC,
  title = {Spin-wave reservoir chips with short-term memory for high-speed estimation of external magnetic fields},
  author = {Nagase, Sho and Nezu, Shoki and Sekiguchi, Koji},
  journal = {Physical Review Applied},
  volume = {22},
  issue = {2},
  pages = {024072},
  numpages = {10},
  year = {2024},
  month = {Aug},
  publisher = {American Physical Society},
  doi = {10.1103/PhysRevApplied.22.024072},
  url = {https://link.aps.org/doi/10.1103/PhysRevApplied.22.024072}
}

@article{spinwave_PRC2,
author = {Namiki, Wataru and Nishioka, Daiki and Yamaguchi, Yu and Tsuchiya, Takashi and Higuchi, Tohru and Terabe, Kazuya},
title = {Experimental Demonstration of High-Performance Physical Reservoir Computing with Nonlinear Interfered Spin Wave Multidetection},
journal = {Advanced Intelligent Systems},
volume = {5},
number = {12},
pages = {2300228},
doi = {https://doi.org/10.1002/aisy.202300228},
url = {https://advanced.onlinelibrary.wiley.com/doi/abs/10.1002/aisy.202300228},
year = {2023}
}

@article{collet2017spin,
  title={Spin-wave propagation in ultra-thin YIG based waveguides},
  author={Collet, Martin and Gladii, Olga and Evelt, Michael and Bessonov, V and Soumah, L and Bortolotti, P and Demokritov, SO and Henry, Y and Cros, V and Bailleul, M and others},
  journal={Applied Physics Letters},
  volume={110},
  number={9},
  year={2017},
  publisher={AIP Publishing}
}

@article{nikolaev2024resonant,
  title={Resonant generation of propagating second-harmonic spin waves in nano-waveguides},
  author={Nikolaev, Kirill O and Lake, SR and Schmidt, G and Demokritov, SO and Demidov, VE},
  journal={Nature Communications},
  volume={15},
  number={1},
  pages={1827},
  year={2024},
  publisher={Nature Publishing Group UK London}
}

@article{spinwave_waveguide,
author = {Nikolaev, Kirill O. and Lake, Stephanie R. and Schmidt, Georg and Demokritov, Sergej O. and Demidov, Vladislav E.},
title = {Zero-Field Spin Waves in YIG Nanowaveguides},
journal = {Nano Letters},
volume = {23},
number = {18},
pages = {8719-8724},
year = {2023},
doi = {10.1021/acs.nanolett.3c02725},
    note ={PMID: 37691265},

URL = { 
        https://doi.org/10.1021/acs.nanolett.3c02725
}}

@article{kubo1966fluctuation,
  title={The fluctuation-dissipation theorem},
  author={Kubo, Rep},
  journal={Reports on progress in physics},
  volume={29},
  number={1},
  pages={255},
  year={1966},
  publisher={IOP Publishing}
}

@article{Einstein_FDT,
author = {Einstein, A.},
title = {Über die von der molekularkinetischen Theorie der Wärme geforderte Bewegung von in ruhenden Flüssigkeiten suspendierten Teilchen},
journal = {Annalen der Physik},
volume = {322},
number = {8},
pages = {549-560},
doi = {https://doi.org/10.1002/andp.19053220806},
url = {https://onlinelibrary.wiley.com/doi/abs/10.1002/andp.19053220806},
year = {1905}
}

@article{FDT1951,
  title = {Irreversibility and Generalized Noise},
  author = {Callen, Herbert B. and Welton, Theodore A.},
  journal = {Phys. Rev.},
  volume = {83},
  issue = {1},
  pages = {34--40},
  numpages = {0},
  year = {1951},
  month = {Jul},
  publisher = {American Physical Society},
  doi = {10.1103/PhysRev.83.34},
  url = {https://link.aps.org/doi/10.1103/PhysRev.83.34}
}

@article{kurikawa2025fluctuation,
  title={Fluctuation-learning relationship in recurrent neural networks},
  author={Kurikawa, Tomoki and Kaneko, Kunihiko},
  journal={Nature Communications},
  volume={16},
  number={1},
  pages={9663},
  year={2025},
  publisher={Nature Publishing Group UK London}
}

@article{Jaeger_2004_esn,
author = {Herbert Jaeger  and Harald Haas },
title = {Harnessing Nonlinearity: Predicting Chaotic Systems and Saving Energy in Wireless Communication},
journal = {Science},
volume = {304},
number = {5667},
pages = {78-80},
year = {2004},
doi = {10.1126/science.1091277},
URL = {https://www.science.org/doi/abs/10.1126/science.1091277}}

@techreport{jaeger2001echo,
  author      = {Herbert Jaeger},
  title       = {The ``Echo State'' Approach to Analysing and Training Recurrent Neural Networks},
  institution = {German National Research Center for Information Technology (GMD)},
  number      = {148},
  year        = {2001}
}

@article{Matteo2021,
author = {Matteo Cucchi  and Christopher Gruener  and Lautaro Petrauskas  and Peter Steiner  and Hsin Tseng  and Axel Fischer  and Bogdan Penkovsky  and Christian Matthus  and Peter Birkholz  and Hans Kleemann  and Karl Leo },
title = {Reservoir computing with biocompatible organic electrochemical networks for brain-inspired biosignal classification},
journal = {Science Advances},
volume = {7},
number = {34},
pages = {eabh0693},
year = {2021},
doi = {10.1126/sciadv.abh0693},
URL = {https://www.science.org/doi/abs/10.1126/sciadv.abh0693}}

@article{pnas2023,
author = {Takuma Sumi  and Hideaki Yamamoto  and Yuichi Katori  and Koki Ito  and Satoshi Moriya  and Tomohiro Konno  and Shigeo Sato  and Ayumi Hirano-Iwata },
title = {Biological neurons act as generalization filters in reservoir computing},
journal = {Proceedings of the National Academy of Sciences},
volume = {120},
number = {25},
pages = {e2217008120},
year = {2023},
doi = {10.1073/pnas.2217008120},
URL = {https://www.pnas.org/doi/abs/10.1073/pnas.2217008120}}

@article{paquot2012optoelectronic,
  title={Optoelectronic reservoir computing},
  author={Paquot, Yvan and Duport, Francois and Smerieri, Antoneo and Dambre, Joni and Schrauwen, Benjamin and Haelterman, Marc and Massar, Serge},
  journal={Scientific reports},
  volume={2},
  number={1},
  pages={287},
  year={2012},
  publisher={Nature Publishing Group UK London}
}

@article{Kalinikos_1986,
doi = {10.1088/0022-3719/19/35/014},
url = {https://doi.org/10.1088/0022-3719/19/35/014},
year = {1986},
month = {dec},
publisher = {},
volume = {19},
number = {35},
pages = {7013},
author = {B A Kalinikos and A N Slavin},
title = {Theory of dipole-exchange spin wave spectrum for ferromagnetic films with mixed exchange boundary conditions},
journal = {Journal of Physics C: Solid State Physics}
}

@article{chumak2015magnon,
  title={Magnon spintronics},
  author={Chumak, Andrii V and Vasyuchka, Vitaliy I and Serga, Alexander A and Hillebrands, Burkard},
  journal={Nature physics},
  volume={11},
  number={6},
  pages={453--461},
  year={2015},
  publisher={Nature Publishing Group UK London}
}

@article{Rezende2009,
  title = {Theory of coherence in Bose-Einstein condensation phenomena in a microwave-driven interacting magnon gas},
  author = {Rezende, Sergio M.},
  journal = {Physical Review B},
  volume = {79},
  issue = {17},
  pages = {174411},
  numpages = {18},
  year = {2009},
  month = {May},
  publisher = {American Physical Society},
  doi = {10.1103/PhysRevB.79.174411},
  url = {https://link.aps.org/doi/10.1103/PhysRevB.79.174411}
}

@article{ushio2023computational,
  title={Computational capability of ecological dynamics},
  author={Ushio, Masayuki and Watanabe, Kazufumi and Fukuda, Yasuhiro and Tokudome, Yuji and Nakajima, Kohei},
  journal={Royal Society open science},
  volume={10},
  number={4},
  pages={221614},
  year={2023},
  publisher={The Royal Society}
}

@article{quantumPRC_2017,
  title = {Harnessing Disordered-Ensemble Quantum Dynamics for Machine Learning},
  author = {Fujii, Keisuke and Nakajima, Kohei},
  journal = {Physical Review Applied},
  volume = {8},
  issue = {2},
  pages = {024030},
  numpages = {20},
  year = {2017},
  month = {Aug},
  publisher = {American Physical Society},
  doi = {10.1103/PhysRevApplied.8.024030},
  url = {https://link.aps.org/doi/10.1103/PhysRevApplied.8.024030}
}

@article{torrejon2017neuromorphic,
  title={Neuromorphic computing with nanoscale spintronic oscillators},
  author={Torrejon, Jacob and Riou, Mathieu and Araujo, Flavio Abreu and Tsunegi, Sumito and Khalsa, Guru and Querlioz, Damien and Bortolotti, Paolo and Cros, Vincent and Yakushiji, Kay and Fukushima, Akio and others},
  journal={Nature},
  volume={547},
  number={7664},
  pages={428--431},
  year={2017},
  publisher={Nature Publishing Group UK London}
}

@article{du2017reservoir,
  title={Reservoir computing using dynamic memristors for temporal information processing},
  author={Du, Chao and Cai, Fuxi and Zidan, Mohammed A and Ma, Wen and Lee, Seung Hwan and Lu, Wei D},
  journal={Nature communications},
  volume={8},
  number={1},
  pages={2204},
  year={2017},
  publisher={Nature Publishing Group UK London}
}

@article{moon2019temporal,
  title={Temporal data classification and forecasting using a memristor-based reservoir computing system},
  author={Moon, John and Ma, Wen and Shin, Jong Hoon and Cai, Fuxi and Du, Chao and Lee, Seung Hwan and Lu, Wei D},
  journal={Nature Electronics},
  volume={2},
  number={10},
  pages={480--487},
  year={2019},
  publisher={Nature Publishing Group UK London}
}

@article{zhong2021dynamic,
  title={Dynamic memristor-based reservoir computing for high-efficiency temporal signal processing},
  author={Zhong, Yanan and Tang, Jianshi and Li, Xinyi and Gao, Bin and Qian, He and Wu, Huaqiang},
  journal={Nature communications},
  volume={12},
  number={1},
  pages={408},
  year={2021},
  publisher={Nature Publishing Group UK London}
}

@article{sakemi2023sparse,
  title={Sparse-firing regularization methods for spiking neural networks with time-to-first-spike coding},
  author={Sakemi, Yusuke and Yamamoto, Kakei and Hosomi, Takeo and Aihara, Kazuyuki},
  journal={Scientific Reports},
  volume={13},
  number={1},
  pages={22897},
  year={2023},
  publisher={Nature Publishing Group UK London}
}

@article{nicola2017supervised,
  title={Supervised learning in spiking neural networks with FORCE training},
  author={Nicola, Wilten and Clopath, Claudia},
  journal={Nature communications},
  volume={8},
  number={1},
  pages={2208},
  year={2017},
  publisher={Nature Publishing Group UK London}
}

@article{xue2013computational,
  title={Computational capability of liquid state machines with spike-timing-dependent plasticity},
  author={Xue, Fangzheng and Hou, Zhicheng and Li, Xiumin},
  journal={Neurocomputing},
  volume={122},
  pages={324--329},
  year={2013},
  publisher={Elsevier}
}

@article{MAASS2004593,
title = {On the computational power of circuits of spiking neurons},
journal = {Journal of Computer and System Sciences},
volume = {69},
number = {4},
pages = {593-616},
year = {2004},
issn = {0022-0000},
doi = {https://doi.org/10.1016/j.jcss.2004.04.001},
url = {https://www.sciencedirect.com/science/article/pii/S0022000004000406},
author = {Wolfgang Maass and Henry Markram}
}

@article{martinez2021dynamical,
  title={Dynamical phase transitions in quantum reservoir computing},
  author={Mart{\'\i}nez-Pe{\~n}a, Rodrigo and Giorgi, Gian Luca and Nokkala, Johannes and Soriano, Miguel C and Zambrini, Roberta},
  journal={Physical Review Letters},
  volume={127},
  number={10},
  pages={100502},
  year={2021},
  publisher={APS}
}

@article{nakajima2019boosting,
  title={Boosting computational power through spatial multiplexing in quantum reservoir computing},
  author={Nakajima, Kohei and Fujii, Keisuke and Negoro, Makoto and Mitarai, Kosuke and Kitagawa, Masahiro},
  journal={Physical Review Applied},
  volume={11},
  number={3},
  pages={034021},
  year={2019},
  publisher={APS}
}

@article{mujal2021opportunities,
  title={Opportunities in quantum reservoir computing and extreme learning machines},
  author={Mujal, Pere and Mart{\'\i}nez-Pe{\~n}a, Rodrigo and Nokkala, Johannes and Garc{\'\i}a-Beni, Jorge and Giorgi, Gian Luca and Soriano, Miguel C and Zambrini, Roberta},
  journal={Advanced Quantum Technologies},
  volume={4},
  number={8},
  pages={2100027},
  year={2021},
  publisher={Wiley Online Library}
}

@article{THIEDE201923,
title = {Gradient based hyperparameter optimization in Echo State Networks},
journal = {Neural Networks},
volume = {115},
pages = {23-29},
year = {2019},
issn = {0893-6080},
doi = {https://doi.org/10.1016/j.neunet.2019.02.001},
url = {https://www.sciencedirect.com/science/article/pii/S0893608019300413},
author = {Luca Anthony Thiede and Ulrich Parlitz}
}

@article{OZTURK2020215,
title = {Optimizing echo state network through a novel fisher maximization based stochastic gradient descent},
journal = {Neurocomputing},
volume = {415},
pages = {215-224},
year = {2020},
issn = {0925-2312},
doi = {https://doi.org/10.1016/j.neucom.2020.07.034},
url = {https://www.sciencedirect.com/science/article/pii/S0925231220311425},
author = {Muhammed Maruf Öztürk and İbrahim Arda Cankaya and Deniz İpekçi}
}

@article{Liu_nakajima2025,
  title = {Exploiting chaotic dynamics as deep neural networks},
  author = {Liu, Shuhong and Akashi, Nozomi and Huang, Qingyao and Kuniyoshi, Yasuo and Nakajima, Kohei},
  journal = {Physical Review Research},
  volume = {7},
  issue = {3},
  pages = {033031},
  numpages = {14},
  year = {2025},
  month = {Jul},
  publisher = {American Physical Society},
  doi = {10.1103/l8dq-mphg},
  url = {https://link.aps.org/doi/10.1103/l8dq-mphg}
}

@article{Tomioka_nakajima2025,
author = {Tomioka, Hiroki and Inoue, Katsuma and Kuniyoshi, Yasuo and Nakajima, Kohei},
title = {Backpropagation Through Soft Body: Investigating Information Processing in Brain–Body Coupling Systems},
journal = {Advanced Robotics Research},
volume = {n/a},
year = {2025},
number = {n/a},
pages = {e202500159},
doi = {https://doi.org/10.1002/adrr.202500159},
url = {https://advanced.onlinelibrary.wiley.com/doi/abs/10.1002/adrr.202500159}
}

@article{tanaka2022continuum,
  title={Continuum-body-pose estimation from partial sensor information using recurrent neural networks},
  author={Tanaka, Kazutoshi and Minami, Yuna and Tokudome, Yuji and Inoue, Katsuma and Kuniyoshi, Yasuo and Nakajima, Kohei},
  journal={IEEE Robotics and Automation Letters},
  volume={7},
  number={4},
  pages={11244--11251},
  year={2022},
  publisher={IEEE}
}

@techreport{jaeger2001short,
  title  = {Short Term Memory in Echo State Networks},
  author = {Jaeger, Herbert},
  year   = {2001},
  institution = {German National Research Center for Information Technology (GMD)}
}

@article{dambre2012information,
  title={Information processing capacity of dynamical systems},
  author={Dambre, Joni and Verstraeten, David and Schrauwen, Benjamin and Massar, Serge},
  journal={Scientific reports},
  volume={2},
  number={1},
  pages={514},
  year={2012},
  publisher={Nature Publishing Group UK London}
}

@article{guan2025noise,
  title={How noise affects memory in linear recurrent networks},
  author={Guan, JingChuan and Kubota, Tomoyuki and Kuniyoshi, Yasuo and Nakajima, Kohei},
  journal={Physical Review Research},
  volume={7},
  number={2},
  pages={023049},
  year={2025},
  publisher={APS}
}

@article{imai2025gradient,
  title={Gradient-based optimization of spintronic devices},
  author={Imai, Yusuke and Liu, Shuhong and Akashi, Nozomi and Nakajima, Kohei},
  journal={Applied Physics Letters},
  volume={126},
  number={8},
  year={2025},
  publisher={AIP Publishing}
}

@article{tsunegi2023information,
  title={Information processing capacity of spintronic oscillator},
  author={Tsunegi, Sumito and Kubota, Tomoyuki and Kamimaki, Akira and Grollier, Julie and Cros, Vincent and Yakushiji, Kay and Fukushima, Akio and Yuasa, Shinji and Kubota, Hitoshi and Nakajima, Kohei and others},
  journal={Advanced Intelligent Systems},
  volume={5},
  number={9},
  pages={2300175},
  year={2023},
  publisher={Wiley Online Library}
}

@article{Kubota2021,
  title = {Unifying framework for information processing in stochastically driven dynamical systems},
  author = {Kubota, Tomoyuki and Takahashi, Hirokazu and Nakajima, Kohei},
  journal = {Phys. Rev. Res.},
  volume = {3},
  issue = {4},
  pages = {043135},
  numpages = {23},
  year = {2021},
  month = {Nov},
  publisher = {American Physical Society},
  doi = {10.1103/PhysRevResearch.3.043135},
  url = {https://link.aps.org/doi/10.1103/PhysRevResearch.3.043135}
}

\end{document}